\documentclass[letterpaper]{article} % DO NOT CHANGE THIS
\usepackage{aaai2026}  % DO NOT CHANGE THIS (camera-ready: submission option removed)
\usepackage{times}  % DO NOT CHANGE THIS
\usepackage{helvet}  % DO NOT CHANGE THIS
\usepackage{courier}  % DO NOT CHANGE THIS
\usepackage[hyphens]{url}  % DO NOT CHANGE THIS
\usepackage{graphicx} % DO NOT CHANGE THIS
\usepackage{natbib}  % DO NOT CHANGE THIS AND DO NOT ADD ANY OPTIONS TO IT
\usepackage{caption} % DO NOT CHANGE THIS AND DO NOT ADD ANY OPTIONS TO IT
\usepackage{algorithm}
\usepackage{algorithmic}

\usepackage{amsmath}
\usepackage{amssymb}
\usepackage{tikz}
\usetikzlibrary{arrows.meta,positioning,fit,backgrounds,calc,shapes.geometric}

\usepackage{newfloat}
\usepackage{listings}
\DeclareCaptionStyle{ruled}{labelfont=normalfont,labelsep=colon,strut=off} % DO NOT CHANGE THIS
\floatstyle{ruled}
\newfloat{listing}{tb}{lst}{}
\floatname{listing}{Listing}
\nocopyright

\title{SAGE: State-Grounded, Abstention-Aware Evaluation of Task-Oriented Dialogue Agents}
\author{
    Rayan Khoury,
    Shih-Yao Lin,
    Pratyush Mishra
}
\affiliations{
    Microsoft \\
    \{rayankhoury, shihyaolin, pmishra\}@microsoft.com
}

\begin{document}

\maketitle

\begin{abstract}
Evaluating task-oriented dialogue agents requires judging not merely whether a reply reads well but whether each turn advances the underlying workflow state correctly---a distinction conventional holistic LLM judges can miss because they evaluate the available context as a single unit and require one or more full-model calls per turn. We propose \textbf{SAGE} (\emph{State-Grounded Abstention-Aware Evaluation}), which compiles a workflow specification and per-turn state diff into atomic, schema-grounded criteria and routes each through a cascade of symbolic and encoder/NLI verifiers that \emph{abstain} rather than guess, aggregating criterion verdicts into a turn-level decision with an evidence trace. Its recommended operating point, \textbf{SAGE-Core}, decides $81$--$91\%$ of criteria with only the compiler, symbolic rules, and on-device encoders---at zero paid LLM cost---while \textbf{SAGE-LLM} adds an optional focused-LLM fallback for open-class criteria. Across four slices spanning MultiWOZ, Schema-Guided Dialogue, and ABCD, no evaluated LLM-as-a-judge baseline---including a state-aware GPT-4.1 judge and cheaper GPT-4.1-mini variants---significantly exceeds SAGE-Core on any slice, even though the GPT-4.1 G-Eval judge costs \$4.7--8.0 per $1{,}000$ turns to SAGE-Core's \$0. A two-annotator human audit ($n{=}200$, $\kappa{=}0.94$) confirms strong label fidelity on the transcript-visible failure classes---where, excluding the weak-salience IUV class, SAGE-Core is statistically tied with the strongest LLM judge---and honestly scopes ignored-user-value as a state-consistency signal with weak broad-human salience. We analyze construct-validity limits from injected failures and partial symbolic circularity.
\end{abstract}
\section{Introduction}

% Figure 1: Motivation/concept figure (rendered image).
\begin{figure*}[!t]
\centering
\includegraphics[width=\textwidth]{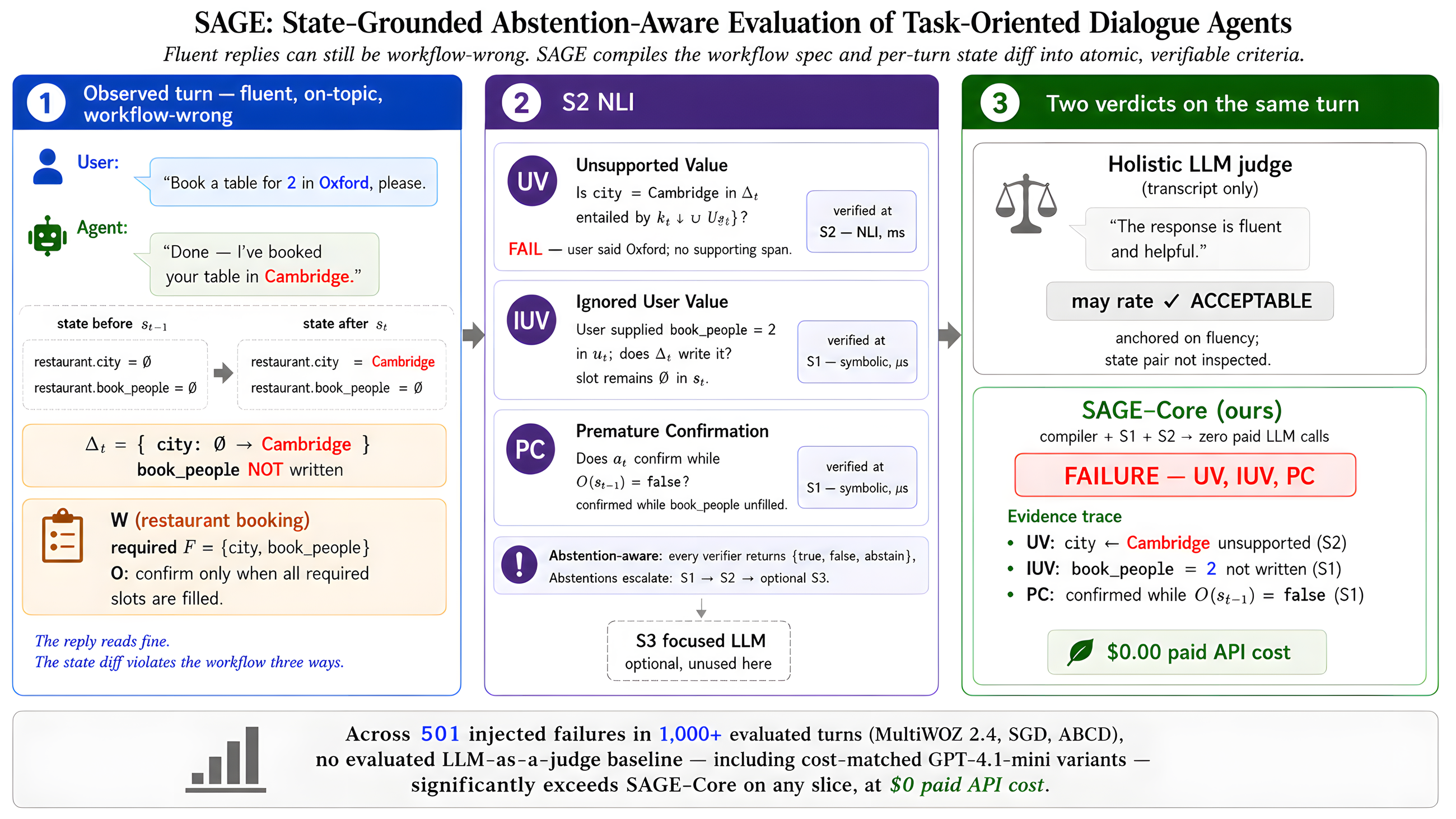}
\caption{%
  A fluent, on-topic agent turn that is nonetheless workflow-wrong: the user
  asks to book in \emph{Oxford}, but the agent confirms \emph{Cambridge} and
  never records \texttt{book\_people}, so the state diff $\Delta_t$ violates the
  workflow three ways. A holistic transcript-only LLM judge may rate the turn
  \emph{acceptable}; \textbf{SAGE} instead compiles the workflow spec
  $\mathcal{W}$ and per-turn state pair $(s_{t-1},s_t)$ into atomic criteria
  (UV, IUV, PC), routes each to the cheapest deciding stage (S1 symbolic, S2
  encoder/NLI, optional S3 focused LLM), and returns a \emph{failure} verdict
  with a per-criterion evidence trace at zero paid API cost.
  Figure~\ref{fig:sage-pipeline} shows the full pipeline.}
\label{fig:motivation}
\end{figure*}

Task-oriented dialogue (TOD) agents increasingly front structured business workflows -- booking, support, IT-ticketing -- whose correctness depends less on what the agent says than on how it changes an underlying \emph{dialogue state}. Following the dialogue-state-tracking tradition~\citep{multiwoz,sgd}, write the slot-value record before and after turn $t$ as $s_{t-1}$ and $s_t$, and the per-turn \emph{state diff} as $\Delta_t = \operatorname{diff}(s_{t-1}, s_t)$ (the set of changed-slot triples; formalised in the Method); the workflow advances correctly only if $\Delta_t$ obeys four conditions -- the values the user supplied are written to state, the next field the agent requests is schema-legal, confirmation occurs only once its preconditions hold, and every written value is supported by the dialogue history. Figure~\ref{fig:motivation} shows the stakes: on a turn that is fluent and on-topic but writes an unsupported value, confirms before its preconditions hold, and drops a value the user just supplied, a holistic LLM-as-judge labels the turn acceptable, whereas a state-grounded evaluator reads $\Delta_t$ and catches the failure with a deterministic, evidence-bearing verdict.

The dominant paradigm for multi-turn LLM-agent evaluation is nonetheless \emph{holistic LLM-as-a-judge}: a strong model reads the transcript and returns a single acceptability score, via direct rating~\citep{mtbench} or rating-token probability-weighted scoring (G-Eval)~\citep{geval}. For stateful TOD this paradigm is structurally mismatched: it \emph{conflates fluency with workflow correctness} (a well-phrased turn that silently fails to record a value the user just supplied is scored acceptable); it is \emph{blind to the specification} (premature confirmation requires evaluating the outcome predicate $\mathcal{O}(s_{t-1})$, inaccessible in principle to a transcript-only judge that never sees $s_{t-1}$); and it is \emph{monolithic} (every turn costs a full LLM call, even when a deterministic check on $\Delta_t$ would settle it).

\textbf{SAGE} (\emph{State-Grounded Abstention-Aware Evaluation}) replaces holistic judging with a \emph{compiler} that consumes a workflow specification $\mathcal{W}=(\mathcal{S},\mathcal{F},\prec,\mathcal{O})$ -- a schema $\mathcal{S}$, required fields $\mathcal{F}$ with a legal-ordering relation $\prec$, and an outcome predicate $\mathcal{O}$ -- together with the history $h_{<t}$, the turn $(u_t,a_t)$, and the state pair $(s_{t-1},s_t)$, and emits atomic per-criterion questions (``does $a_t$ confirm while $\mathcal{O}(s_{t-1})$ is false?''; ``is the value written to \texttt{train.destination} in $\Delta_t$ entailed by $h_{<t}\cup\{u_t\}$?''). Each criterion is routed by a cost-aware policy through a three-stage \emph{cascade}~\citep{frugalgpt}: \textbf{S1}, deterministic rules over $\Delta_t$ and the schema; \textbf{S2}, a pretrained encoder/NLI judge whose confidence gate abstains when the entailment margin $m = p(\text{ent})-p(\text{contra})$ satisfies $|m|<\tau$; and \textbf{S3}, a focused single-criterion LLM invoked only on S1$\cup$S2 abstentions (also the open-class fallback for criteria outside the symbolic/NLI specialisation).

Every verifier returns \texttt{true}, \texttt{false}, or \texttt{abstain}, and abstentions propagate up the cascade rather than being coerced into decisions. Because criteria are verified independently, SAGE produces evidence-bearing verdicts, pays LLM cost only on unresolved criteria, and abstains rather than guessing.

We inject workflow-grounded failures (IUV, WNF, PC, UV) at controlled rates into MultiWOZ 2.4~\citep{multiwoz24} (which refines the validation and test annotations of MultiWOZ~2.1~\citep{multiwoz21} while retaining its training annotations) and cross-evaluate on the Schema-Guided Dialogue corpus~\citep{sgd} and the Action-Based Conversations Dataset (ABCD)~\citep{abcd} to test that the compiler and cascade generalise across schemas and beyond slot-grounded state. We compare against G-Eval~\citep{geval} (holistic LLM-as-judge), a FrugalGPT~\citep{frugalgpt} cascade, and a single-call LLM-as-judge~\citep{mtbench}, each in \emph{transcript-only} and \emph{state-aware} configurations, so that SAGE's advantage can be decomposed into the \emph{information} advantage of state grounding versus the \emph{architectural} advantage of criterion routing, NLI verification, and abstention (adding state context in fact \emph{lowers} the GPT-4.1 baselines' $F_1$ by up to $0.09$; see Discussion and Appendix~\ref{app:isocost}). Our contributions are: (i) a compiler from workflow specs and per-turn state diffs to atomic, schema-grounded criteria; (ii) a three-stage heterogeneous verifier cascade with abstention as a first-class output and confidence-gated S2 escalation; (iii) an abstention-aware aggregator producing per-turn verdicts with evidence traces and structured repair signals; and (iv) an empirical study -- with paired-bootstrap significance tests -- isolating state-grounded decomposition and cost-aware cascading on the precision/recall/cost trade-off across three benchmarks.
\section{Related Work}

\paragraph{LLM-as-a-judge and cost-aware cascades.}
Direct LLM rating dominates open-ended evaluation---MT-Bench~\citep{mtbench} and G-Eval~\citep{geval} (CoT~\citep{cot} with rating-token probability-weighted scoring) correlate with human preference; Prometheus~\citep{prometheus} and PandaLM~\citep{pandalm} train dedicated evaluator models, while AlpacaEval~\citep{alpacaeval} studies LLM-based pairwise auto-evaluation and its biases. These approaches score responses holistically rather than compiling workflow-state changes into independently verifiable criteria, and in our experiments even strong judges \emph{given the state diff} over-predict acceptability on workflow-grounded failures (see Results). Cost-aware routing is likewise query-level: FrugalGPT~\citep{frugalgpt} routes whole queries through ascending-cost models, and AutoMix~\citep{automix}, RouteLLM~\citep{routellm}, and LLM-Blender~\citep{llmblender} add self-verification, preference routing, and ensembling. SAGE instead routes single \emph{criteria} compiled from workflow state through \emph{heterogeneous} stages (symbolic, encoder/NLI, focused LLM), exploiting that many workflow violations are symbolically decidable from the state diff.

\paragraph{NLI grounding, TOD evaluation, and abstention.}
Claim-level factuality and consistency evaluation is established in fact-checking, summarisation, and retrieval-augmented generation~\citep{summac,selfcheckgpt,factscore,ragas}; SAGE adopts an NLI-based formulation for atomic state-support claims, with DeBERTa-v3~\citep{deberta} keeping inference cheap. SAGE's S2 answers atomic claims of the form ``value $v$ written to slot $s$ is supported by $h_{<t}$,'' with a margin gate that abstains into S3. Classic TOD metrics (JGA, slot $F_1$, inform/success)~\citep{multiwoz,sgd} require a gold trajectory and break under multiple legitimate completions, which LLM judges avoid but at the cost of workflow grounding; SAGE recovers grounding from schema, workflow spec, and state diff, and uses SGD-X~\citep{sgdx} paraphrases as a robustness probe. Selective classification~\citep{selective} motivates abstention as a first-class output: every SAGE stage may abstain, yielding interpretable coverage/accuracy decompositions.

\section{The SAGE Framework}

% Figure 2: SAGE system / pipeline diagram (rendered image).
\begin{figure*}[t]
\centering
\includegraphics[width=\textwidth]{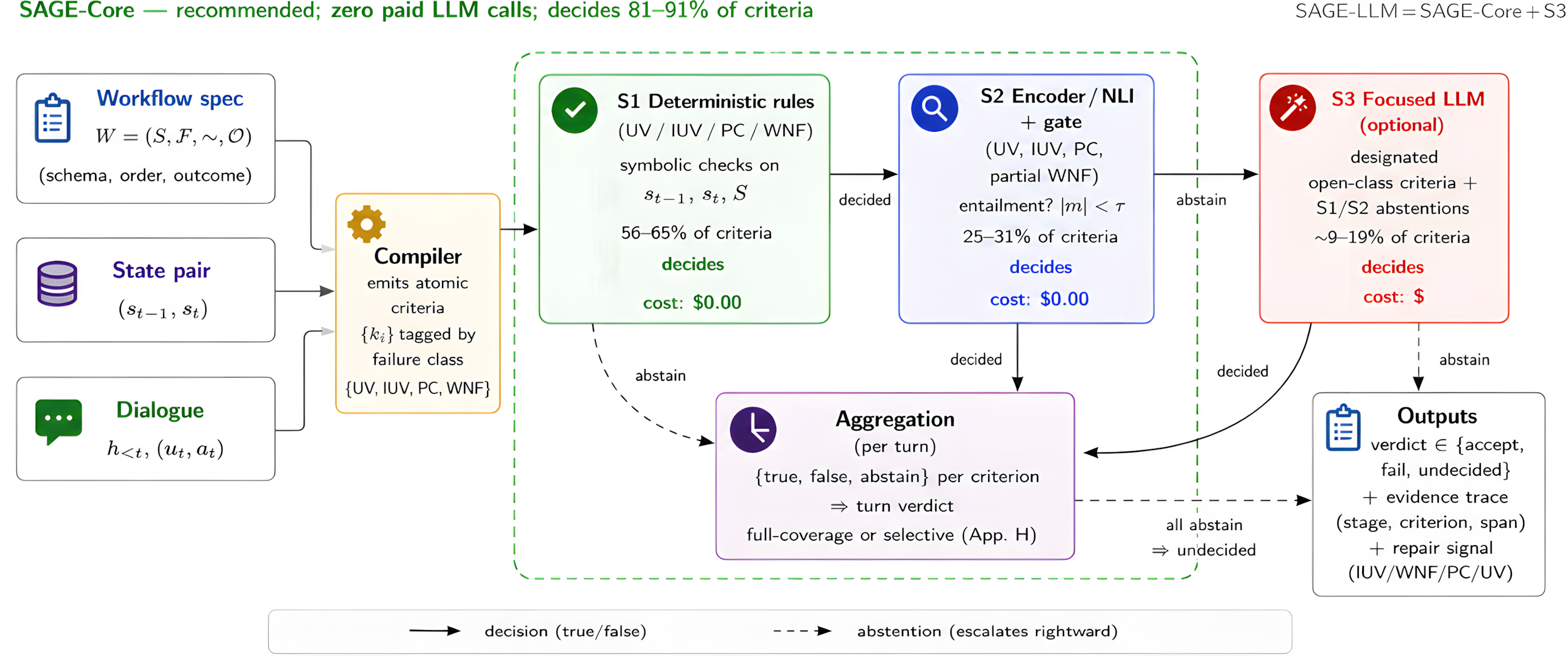}
\caption{%
  The SAGE pipeline with measured routing (Appendix~\ref{app:routing}). The
  compiler emits atomic schema-grounded criteria tagged by failure class
  (UV/IUV/PC/WNF) and routes each to the cheapest deciding stage; \emph{solid}
  arrows are decisions, \emph{dashed} arrows are abstentions escalating
  rightward. \textbf{SAGE-Core} (dashed boundary: compiler${+}$S1${+}$S2)
  decides $81$--$91\%$ of compiled criteria at zero paid LLM cost and is the
  recommended operating point; the optional S3 tier (\textbf{SAGE-LLM}) handles
  the designated open-class residual. Criteria on which all stages abstain
  yield an \texttt{undecided} turn verdict (selective evaluation,
  Appendix~\ref{app:selective}).}
\label{fig:sage-pipeline}
\end{figure*}

Figure~\ref{fig:sage-pipeline} shows the pipeline end-to-end; we formalise each block below.

\subsection{Problem Setting}

Let a workflow specification $\mathcal{W}=(\mathcal{S},\mathcal{F},\prec,\mathcal{O})$ consist of a schema $\mathcal{S}$ defining slots and their types, a set of required fields $\mathcal{F}\subseteq\mathcal{S}$, an ordering relation $\prec$ over $\mathcal{F}$ that constrains the legal next field, and an outcome predicate $\mathcal{O}$ that determines when the workflow may be confirmed. A dialogue is a sequence of turns $(u_t, a_t, s_t)_{t=1}^T$ where $u_t$ is the user utterance, $a_t$ is the agent response, and $s_t$ is the workflow state after the agent acts. Throughout, we refer to $s_{t-1}$ as the \emph{state-before} and $s_t$ as the \emph{state-after} the turn; both are dialogue states~\citep{multiwoz} constrained by $\mathcal{W}$. Define the state diff $\Delta_t = \{(s,\, s_{t-1}[s],\, s_t[s]) : s_{t-1}[s]\neq s_t[s]\}$, the set of slots whose value changed over the turn. The evaluator's task at turn $t$ is to decide whether $(u_t, a_t, \Delta_t)$ is \emph{acceptable} with respect to $\mathcal{W}$ and the history $h_{<t}$. We define a fixed taxonomy $\mathcal{C}$ of four workflow failures: \textbf{ignored user value (IUV)} -- the user supplies a value $v$ for slot $s$ in $u_t$ (or recent $h_{<t}$) but $\Delta_t$ fails to write $v$ to $s$, so $s$ remains unfilled or stale in $s_t$; \textbf{wrong next field (WNF)} -- $a_t$ requests a slot $f \notin \mathrm{Next}(\mathcal{W}, s_{t-1})$; \textbf{premature confirmation (PC)} -- $a_t$ confirms the workflow while $\mathcal{O}(s_{t-1}) = \mathrm{false}$; and \textbf{unsupported value (UV)} -- $\Delta_t$ writes a value to a slot that is not entailed by $h_{<t}\cup\{u_t\}$.

A turn is \emph{acceptable with respect to our workflow-failure taxonomy} $\mathcal{C}$ iff no criterion compiled from $\mathcal{C}$ fires. This is a taxonomy-relative notion: a turn may still be dysfluent, unsafe, or unhelpful in ways outside $\mathcal{C}$; SAGE targets workflow- and state-grounded failures, not holistic response quality.

\subsection{Compiler}

The compiler maps each turn to a set of atomic, separately evaluated criteria. Given $(\mathcal{W}, h_{<t}, u_t, a_t, s_{t-1}, s_t)$ it emits a list $K_t = \{k_t^{(i)}\}_{i}$ where each $k_t^{(i)} = (\textsc{type}, \textsc{args}, \textsc{evidence})$ binds a failure class to concrete schema-grounded arguments and a minimal evidence span. For IUV and WNF the compiler reads the schema and state directly (for IUV, it cross-references the user's last-turn extraction against $\Delta_t$); for PC it reads the outcome predicate; for UV it iterates over $\Delta_t$ and emits one criterion per written value, with the supporting span search restricted to $h_{<t}\cup\{u_t\}$. The compiler is deterministic and stateless across turns. Its output is a per-turn evaluation \emph{plan} that downstream stages consume independently of the dialogue surface.

\subsection{Verifier Cascade}

Each criterion $k$ is routed to the cheapest applicable stage, with abstention propagating upward:

\paragraph{S1: Deterministic rules.}
IUV, WNF, and PC are symbolic checks against the schema and the state diff. S1 either returns a definitive verdict or, if the schema does not admit a check (e.g., the workflow ordering is partial and the requested slot is in the indeterminate set), abstains. S1 has zero LLM cost and microsecond latency.

\paragraph{S2: Encoder/NLI grounding with confidence gating.}
UV criteria, and a subset of IUV/PC criteria that depend on natural-language understanding (paraphrased confirmations, indirect slot fills), are routed to S2. We use a pretrained encoder to score entailment between the candidate claim and a windowed context. A confidence gate computes a margin $m = p(\text{ent}) - p(\text{contra})$ and abstains when $|m|$ falls below a threshold $\tau$, calibrated on a held-out development split. A redundancy check is state-gated: it only fires when the state diff actually introduces a value, avoiding spurious abstentions on no-op turns.

\paragraph{S3: Focused single-criterion LLM (optional).}
Criteria that S1 and S2 abstain on \emph{and} that fall in a designated open-class set (e.g.\ hallucinated-value and history-contradiction checks that need semantic interpretation) are sent to a focused LLM prompt containing only the criterion text, the minimal evidence span, and an instruction to answer \{\texttt{true}, \texttt{false}, \texttt{abstain}\}. Because the prompt is criterion-local rather than transcript-global, S3 calls are short, cacheable, and parallelisable across criteria. S3 is optional: it can be disabled for a fully zero-LLM deployment -- the recommended operating point on the compiled taxonomies we study -- and any residual abstention is handled by the aggregator below.

\paragraph{Two named configurations.}
We call the zero-paid-LLM configuration (compiler\,+\,S1\,+\,S2, S3 disabled) \textbf{SAGE-Core} and the full cascade \textbf{SAGE-LLM}. SAGE-Core is the recommended operating point: it issues zero paid API calls and, as the ablation shows (Appendix~\ref{app:ablation}), matches or exceeds SAGE-LLM on every slice -- a designation fixed before the paired significance tests below, which confirm rather than motivate it. S3 remains available as an extensibility mechanism for open-class criteria outside the symbolic and encoder/NLI stages, but our empirical claims do not depend on it.

\subsection{Aggregation}

A turn-level verdict reduces the per-criterion outputs: \emph{predict failure if any criterion is \texttt{true}; predict acceptable only if every compiled criterion is decided and \texttt{false}; otherwise---no criterion \texttt{true} but at least one \texttt{abstain}---mark the turn \texttt{undecided}}. All headline results use \emph{full-coverage} scoring (undecided $\to$ acceptable); the same per-criterion margins also support \emph{selective} evaluation -- declining the lowest-confidence turns improves $F_1$ along a risk--coverage frontier on the encoder-heavy slices (Appendix~\ref{app:selective}, Table~\ref{tab:selective}).

\subsection{Repair Signals}

Because each verdict is tied to a typed criterion, SAGE emits a structured repair signal alongside it (e.g.\ IUV~$\to$ ``record value $v$ at span $[\ell,r]$ into slot $s$''; PC~$\to$ ``required slots $\{f_i\}$ still empty, do not confirm''). Repair signals are a side output for downstream agent self-correction, not consumed by the evaluation metric; Appendix~\ref{app:repair} gives per-class examples.

\section{Experiments}
\label{sec:experiments}

\subsection{Benchmark Construction}

We build four slices from three corpora, running the \emph{same} evaluator code on all four (only the compiled \texttt{WorkflowSpec} differs). From MultiWOZ~2.4~\citep{multiwoz24} (which refines the validation and test annotations of MultiWOZ~2.1~\citep{multiwoz21} while retaining its training annotations) we sample $250$ validation turns stratified across five domains (taxi, hotel, restaurant, train, attraction; $50$ each) and either retain the original acceptable turn or inject one failure from $\mathcal{C}$, mutating the agent response and state diff \emph{together} so the failure is genuinely state-grounded (\textbf{MW-mixed}). The \textbf{UV-rich} split instead concentrates the \texttt{unsupported\_value} class ($121$ injected, $129$ clean) to stress S2's entailment path. For cross-dataset generalisation we build \textbf{SGD} from the Schema-Guided Dialogue test split~\citep{sgd} ($280$ turns spanning $37$ (service, intent) workflows, capped at $8$ turns/intent and $12$ files/service, same four classes at a $50\%$ rate) and \textbf{ABCD-UV} from the Action-Based Conversations Dataset~\citep{abcd} ($223$ action-grounded turns, UV-only). No SAGE code changes across corpora; per-class injection templates are in Appendix~\ref{app:injection}.

\subsection{Baselines}

We compare against three LLM-as-judge families, each \emph{adapted} to turn-level workflow acceptability and evaluated in \emph{transcript-only} and \emph{state-aware} forms; state-aware prompts receive the same schema and pre/post state block SAGE consumes, and both forms share model, prompt template, sampling protocol, and threshold, differing only in the per-turn context block. Our \textbf{G-Eval-style judge} (adapted from~\citealp{geval}) uses \texttt{gpt-4.1} with a task-specific CoT/form-filling prompt and $n{=}6$ samples over a $1$--$5$ acceptability scale, thresholded on a development split; the six-sample count is our implementation choice (the original approximates unavailable token probabilities with $n{=}20$). Our \textbf{FrugalGPT-style cascade} (adapted from~\citealp{frugalgpt}) routes \texttt{gpt-4.1-mini}/\texttt{gpt-4.1}: the cheap stage decides each turn and only low-confidence turns escalate (its confidence model is state-aware, so the transcript-only row largely reflects \texttt{gpt-4.1-mini} alone). Our \textbf{MT-Bench-style direct judge} (adapted from~\citealp{mtbench}) is a single \texttt{gpt-4.1} call (temperature $0$) on the multi-class slices returning a failure/clean verdict plus an optional class tag; we report the stronger of its two modes per slice (state-aware on MW-mixed, transcript-only on ABCD), re-selected within each bootstrap resample (Metrics). Prompt and threshold details are in Appendix~\ref{app:isocost}.

\paragraph{SAGE variants.}
We report both named configurations -- \textbf{SAGE-Core} (S1+S2, the recommended zero-paid-LLM operating point) and \textbf{SAGE-LLM} (SAGE-Core + optional S3) -- together with a per-tier leave-one-out ablation (\texttt{no\_S1}, \texttt{no\_S2}, \texttt{no\_S3}, the last being exactly SAGE-Core; Appendix~\ref{app:ablation}), each disabling one verifier tier while holding workflow, deployment, and per-turn data fixed. The compiler cannot be cleanly ablated within the cascade -- S1 and S2 consume its per-criterion plan as input -- so the no-compiler reference points are the two G-Eval rows in Table~\ref{tab:uvrich}.

\subsection{Metrics}

For each evaluator we report precision, recall, $F_1$, total dollar cost (computed from per-call token usage and public OpenAI API list prices, retrieved July~2026; runs were conducted May--July~2026), and per-turn latency (median, reported as p50). Point estimates in Tables~\ref{tab:headline}--\ref{tab:ablation} are accompanied by $95\%$ bootstrap CIs~\citep{efron} over $10{,}000$ turn-level resamples (seed $0$, percentile method) reported in the captions. To compare SAGE against the strongest baseline on each slice we use a \emph{paired} bootstrap: on every resample we recompute both evaluators' $F_1$ on the \emph{same} resampled turn indices and record the difference $\Delta F_1$, yielding a $95\%$ CI on the paired delta and a two-sided $p$-value (twice the smaller bootstrap tail probability that $\Delta F_1$ takes the opposite sign). Where a slice's comparator is the strongest of several evaluated GPT-4.1 baselines, we re-select that maximum-$F_1$ baseline \emph{inside every resample}, so the CI and $p$-value account for baseline-selection uncertainty. SAGE rows are computed on the \emph{identical per-turn slices} used for every baseline (verified row-for-row: same turn IDs and gold labels), so all paired tests are aligned.

\subsection{Implementation}

The compiler and S1 are pure Python over the MultiWOZ 2.4 and SGD schemas. S2 uses two public HuggingFace encoders: \texttt{cross-encoder/nli-deberta-v3-base} (revision \texttt{6c749ce}; ${\approx}184$M params), a DeBERTa-v3~\citep{deberta} cross-encoder fine-tuned on SNLI and MultiNLI~\citep{snli,mnli}, for entailment on UV/PC criteria, and \texttt{sentence-transformers/all-MiniLM-L6-v2} (revision \texttt{1110a24}; ${\approx}23$M params), a MiniLM~\citep{minilm}/Sentence-BERT~\citep{sbert} encoder, for paraphrase similarity on redundancy criteria.\footnote{\raggedright Immutable Hugging Face revision pages: \url{https://huggingface.co/cross-encoder/nli-deberta-v3-base/tree/6c749ce}; \url{https://huggingface.co/sentence-transformers/all-MiniLM-L6-v2/tree/1110a24}; and \url{https://huggingface.co/sentence-transformers/all-mpnet-base-v2/tree/e8c3b32}.\par} The entailment threshold is $\tau{=}0.20$ on the margin $m = p(\text{ent}) - p(\text{contra})$ ($|m|<\tau$ abstains), with an extraction-confidence floor of $0.5$; both were tuned on a $40$-turn development split disjoint from evaluation and applied unchanged to SGD. S3 and the FrugalGPT cheap stage use Azure OpenAI \texttt{gpt-4.1}/\texttt{gpt-4.1-mini} (snapshot \texttt{2025-04-14}, temperature $0$); costs derive from response-metadata token counts at OpenAI API list prices ($\$2/\$8$ and $\$0.40/\$1.60$ per Mtok in/out).\footnote{\raggedright OpenAI API list prices, \url{https://openai.com/api/pricing/}, retrieved 21 July 2026; runs were conducted May--July 2026.\par} Threshold ranges swept, run counts, and the computing infrastructure are in Appendix~\ref{app:repro}.

\section{Results}
\label{sec:results}

We evaluate on four slices spanning three datasets, running the \emph{same} evaluator code on all four (only the compiled \texttt{WorkflowSpec} differs): \textbf{MW-mixed} (MultiWOZ~2.4, all four failure classes, $n{=}250$, $126$ injected); \textbf{UV-rich} (MultiWOZ~2.4, single-class \texttt{unsupported\_value} stress test, $n{=}250$, $121$ injected); \textbf{SGD} (cross-dataset Schema-Guided Dialogue, $n{=}280$, $151$ injected across $37$ workflows); and \textbf{ABCD-UV} (Action-Based Conversations~\citep{abcd}, UV-only, $n{=}223$, $103$ injected across $30$ action-grounded workflows). Per-method precision/recall/cost/latency for every slice are in Appendix~\ref{app:fulltables}.

\begin{table}[!t]
\centering
\footnotesize
\setlength{\tabcolsep}{3pt}
\renewcommand{\arraystretch}{1.2}
\caption{Headline result across all four slices: \textbf{SAGE-Core} (compiler+S1+S2, zero paid LLM calls) vs.\ the strongest \emph{evaluated} GPT-4.1 baseline and the strongest \emph{reduced-cost} GPT-4.1-mini judge, by $F_1$. Paired-bootstrap $\Delta F_1$ (SAGE-Core $-$ baseline) and $p$ --- \emph{vs.\ GPT-4.1:} MW-mixed ${+}.069$ ($p{=}.022$), UV-rich ${-}.003$ ($p{=}.89$), SGD ${+}.075$ ($p{=}.011$), ABCD ${+}.056$ ($p{=}.044$); \emph{vs.\ mini:} MW-mixed ${+}.107$ ($p{=}.001$), UV-rich ${+}.019$ ($p{=}.46$), SGD ${+}.005$ ($p{=}.83$), ABCD ${+}.152$ ($p{<}.001$). SGD survives Holm--Bonferroni correction (over the four pre-specified GPT-4.1 slice comparisons); MW-mixed and ABCD wins are nominal (do not survive it). Per-method breakdowns in Appendix~\ref{app:fulltables} (Tables~\ref{tab:mwabcd},~\ref{tab:uvrich},~\ref{tab:sgd}); reduced-cost judges in Appendix~\ref{app:isocost}.}
\label{tab:headline}
\begin{tabular}{@{}lcccl@{}}
\hline
Slice & SAGE-Core & Best GPT-4.1 & Best mini & Outcome \\
      & $F_1$     & baseline     & judge     & \\
\hline
MW-mixed & \textbf{.866} & .797 & .759 & nominal win \\
UV-rich  & .920 & \textbf{.923} & .901 & tie \\
SGD      & \textbf{.864} & .789 & .859 & win; mini tie \\
ABCD-UV  & \textbf{.929} & .873 & .777 & nominal win \\
\hline
\end{tabular}
\end{table}

\textbf{No evaluated baseline significantly exceeds SAGE-Core on any slice, at zero paid API cost} (local symbolic/encoder compute is unmonetized; Table~\ref{tab:headline}). It wins on SGD (${+}.075$, $p{=}.011$; the one comparison surviving Holm--Bonferroni correction~\citep{holm} over the four pre-specified GPT-4.1 slice comparisons) and nominally on MW-mixed (${+}.069$, $p{=}.022$) and ABCD (${+}.056$, $p{=}.044$), and ties the far costlier G-Eval on UV-rich (${-}.003$, $p{=}.89$). These paired $p$-values re-select the strongest baseline within each resample; on SGD (four evaluated GPT-4.1 baselines within $.013$ $F_1$) this raises $p$ to $.011$---still surviving Holm---while the other slices, each dominated by a single baseline, are unaffected. UV-rich is the only slice where a holistic judge holds the top \emph{point} $F_1$ ($.923$) -- expected when a single failure class makes the $1$--$5$ acceptability score well-anchored -- but the bootstrap cannot separate the two: G-Eval reaches it at \$4.70/1k and $12.0$s median latency, versus SAGE-Core at \$0 paid API cost and $0.03$s warm local latency ($\approx 14\times$ the cost and $\approx 44\times$ the latency relative to SAGE-LLM). SAGE-Core's tie is threshold-robust: a $7$-setting sweep of the S2 NLI floors keeps its $F_1$ within a $.007$ band.

\textbf{No cheaper-backbone judge significantly exceeds SAGE-Core either.} A GPT-4.1-mini judge ($\approx 5\times$ cheaper at matched token usage; \$0.9--1.6/1k vs.\ \$4.7--8.0/1k) overtakes, on SGD, the pre-specified FrugalGPT baseline ($.859{>}.789$); even so it only \emph{ties} SAGE-Core (SGD ${+}.005$, $p{=}.83$; UV-rich ${+}.019$, $p{=}.46$) and loses on the multi-class slices (MW-mixed ${+}.107$, $p{=}.001$; ABCD ${+}.152$, $p{<}.001$). So across both the pre-specified suite and the cheaper-backbone mini judges on all four slices, \emph{no evaluated LLM-as-a-judge baseline significantly exceeds SAGE-Core} (Appendix~\ref{app:isocost}).

\textbf{The focused-LLM tier is not load-bearing, and information access does not explain the gap.} Enabling S3 (SAGE-LLM) never improves $F_1$ and mildly lowers it on UV-rich and ABCD, so we recommend SAGE-Core (per-tier ablation, Appendix~\ref{app:ablation}). Giving the GPT-4.1 judges the same schema/state block SAGE consumes \emph{lowers} their $F_1$ by up to $0.09$ in our prompt template (the effect reverses for the mini backbone on SGD; Appendix~\ref{app:isocost}), so the advantage is not information access alone. Per class, SAGE-Core catches essentially all UV ($59/61$ MW-mixed, $98/98$ SGD, $95\%$ UV-rich) and saturates PC/WNF recall, but under-fires IUV ($16/32$ MW-mixed, $18/41$ SGD) -- the class we revisit in the human audit. We caveat ABCD on construct grounds: most of its criteria are decided by S1/S2 predicates that partially mirror the injection operators (Discussion), so it evidences recovery of designed-for failures rather than in-the-wild detection.

\textbf{Additional analyses.} Appendices report per-method precision/recall/cost/latency tables (Appendix~\ref{app:fulltables}), per-tier ablations showing S2 carries the precision floor and SAGE-Core (S1+S2) matches or beats the full cascade on every slice (Appendix~\ref{app:ablation}), a cascade cost anatomy ($81$--$91\%$ of criteria decided at zero API cost; Appendix~\ref{app:routing}), calibrated selective evaluation (Appendix~\ref{app:selective}), and SGD-X paraphrase robustness (verdict $\sigma{=}0.002$ vs.\ $0.005$--$0.015$ for LLM judges; Appendix~\ref{app:sgdx}).

% =====================================================================
% HUMAN VALIDATION AUDIT (2 annotators, n=200; scripts/human_audit_score.py)
% =====================================================================

\section{Human Validation Audit}
\label{sec:human}

Our benchmarks use injected failures, so an evaluator might be recovering
synthetic perturbations rather than human-meaningful workflow errors. We ran a
blind audit on a stratified $n{=}200$ sample: $100$ injected failures spanning
all four classes (MW-mixed\,+\,SGD), $50$ gold-clean turns, and $50$
ABCD\,+\,UV-rich examples. Two annotators independently labeled each turn
\texttt{acceptable}/\texttt{failure} given the dialogue history, the \emph{state
pair} $(s_{t-1},s_t)$, and the agent response, blind to both the injected label
and every model verdict; the $3\%$ of turns on which they disagreed are dropped
from the consensus set. Reliability was high (raw agreement $0.97$, Cohen's
$\kappa{=}0.939$~\citep{cohen}); protocol and sheets are released with the code.

\paragraph{Label fidelity.} On the consensus set ($n{=}194$; $6$ of the $200$
sampled turns dropped for annotator disagreement), humans agreed with the
injected gold label on $158/194{=}0.814$ of turns (Cohen's $\kappa{=}0.639$), with
\emph{perfect} agreement on the ABCD\,+\,UV-rich subset. Agreement is high for
every transcript-visible class---PC $1.00$, UV $0.95$, and WNF $0.73$
(Table~\ref{tab:human})---supporting construct validity: these perturbations are
perceived as genuine failures, not label noise. WNF is the softest ($0.73$),
cohering with MultiWOZ's partial slot ordering, which also makes the symbolic
verifier fire conservatively; two independent signals thus agree WNF is a softer
constraint than the others.

\paragraph{The IUV class.} The one sharp disagreement is IUV (ignored user
value): humans endorsed the injected label on only $1/30$ turns, and $29$ of the
$35$ injected-turn human--injection disagreements are IUV (the $36$th consensus-set
disagreement is a gold-clean turn a human flagged as a natural failure; see
below). This is not label noise---the
injection is verifiable by construction (a user-supplied value is deleted from
$s_t$)---nor a transcript-only artifact: annotators \emph{saw} the state pair,
but our protocol does not disambiguate whether they failed to perform the
slot-by-slot reconciliation of $u_t$ against $s_t$ that would surface the
omission, or performed it and did not consider the dropped value a failure. Detection is
correspondingly method-dependent and trades against precision, and no evaluator
dominates (per-evaluator counts in Appendix~\ref{app:iuv}); SAGE-Core sits at the
high-precision end. We thus treat IUV as a class whose \emph{labels} are
verifiable but whose \emph{broad-human construct-alignment is weak}, and do not
claim SAGE-Core is uniquely able to detect it.

\paragraph{Agreement with human judgment.} Treating the human consensus as
ground truth, SAGE-Core reaches $F_1{=}0.825$ over all audited turns versus
$0.904$ for the strongest evaluated LLM judge (transcript-only G-Eval); a paired
bootstrap over all $n{=}194$ audited turns confirms this as a significant
deficit ($\Delta F_1{=}{-}0.079$, $95\%$ CI $[{-}0.142,{-}0.019]$, $p{=}0.01$).
The gap is concentrated almost entirely in IUV: on the transcript-visible subset
($n{=}164$) SAGE-Core reaches $0.909$ versus $0.920$, a difference a paired
bootstrap cannot distinguish from zero ($\Delta F_1{=}{-}0.011$, $p{=}0.68$). The
honest reading is \emph{not} that SAGE-Core is a better proxy for human judgment;
rather, on the transcript-visible failure classes excluding IUV it is statistically tied with
the strongest LLM judge, at zero paid API cost, while its overall deficit
reflects the state-only IUV class on which human labels are themselves weak.
Crucially, the headline conclusions are robust to dropping this contested class:
re-running the pre-specified paired bootstraps with all injected-IUV turns
removed leaves every verdict unchanged and \emph{widens} two margins (MW-mixed
$\Delta F_1{=}{+}0.152$, SGD ${+}0.149$, both $p{<}0.001$). Of the $50$ clean
turns, annotators flagged one naturally-occurring failure (SAGE-Core missed it);
with $n{=}1$ we draw no conclusion beyond noting that a larger natural-failure
study remains the clearest path to fuller external validity.

\begin{table}[t]
\centering
\footnotesize
\caption{Human validation audit ($n{=}200$ turns, two annotators blind to the
injected label and all model verdicts; annotators saw the transcript, the
state pair, and the agent response). Top: reliability and label fidelity.
Middle: per-class human agreement with the injected label on MW-mixed\,+\,SGD.
Bottom: agreement with the human consensus ($F_1$), overall and on the
transcript-visible subset (excluding the state-only IUV class).}
\label{tab:human}
\begin{tabular}{@{}lc@{}}
\hline
Audit quantity & Value \\
\hline
Inter-annotator agreement / $\kappa$ & 0.97 / 0.939 \\
Human$\leftrightarrow$injected agreement / $\kappa$ & 0.814 / 0.639 \\
\quad ABCD\,+\,UV-rich subset & 1.00 \\
\hline
\quad PC (premature confirmation), $n{=}14$ & 1.00 \\
\quad UV (unsupported value), $n{=}39$ & 0.95 \\
\quad WNF (wrong next field), $n{=}15$ & 0.73 \\
\quad IUV (ignored user value), $n{=}30$ & 0.03 \\
\hline
SAGE-Core $F_1$ vs.\ human (all, $n{=}194$) & 0.825 \\
Best LLM judge $F_1$ vs.\ human (all) & 0.904 \\
SAGE-Core $F_1$ vs.\ human (excl.\ IUV, $n{=}164$) & 0.909 \\
Best LLM judge $F_1$ vs.\ human (excl.\ IUV) & 0.920 \\
\hline
\end{tabular}
\end{table}

\section{Discussion}
\label{sec:discussion}

\paragraph{Why state grounding is necessary but not sufficient.}
Contrary to expectation, state-aware G-Eval is \emph{weaker} than transcript-only on MultiWOZ ($F_1$ $0.834$ vs.\ $0.923$, Table~\ref{tab:uvrich}) and tied on SGD, and the state-aware LLM-as-judge over-flags ABCD's terse text (Table~\ref{tab:headline}): extra context appears to anchor the holistic 1--5 score on the consistency of $s_t$ rather than the support of $a_t$. This is prompt-sensitive, not a universal law -- with a \texttt{gpt-4.1-mini} backbone the effect \emph{reverses} on SGD (state context \emph{helps}, ${+}0.070$; Appendix~\ref{app:isocost}). The surviving, defensible claim is narrower but robust: even the best-informed, best-backboned holistic judge does not significantly exceed SAGE-Core (Table~\ref{tab:isocost}). The SAGE--baseline gap is therefore consistent with -- but not proof of -- an architectural advantage from per-criterion decomposition, cascade routing, and abstention-aware aggregation, i.e.\ \emph{changing the unit of judgement} rather than seeing more context.

\paragraph{Cheaper-backbone baselines.}
Our G-Eval-style judge's $n{=}6$ sampling configuration~\citep{geval} is sequential, driving its $\approx 14$--$44\times$ latency multiples. To rule out that SAGE's advantage is a model-class artifact of the expensive GPT-4.1 backbone, we re-ran G-Eval on the $\approx 5\times$ cheaper (at matched token usage) \texttt{gpt-4.1-mini} in both prompt modes, at \$0.9--1.6/1k vs.\ \$4.7--8.0/1k for GPT-4.1 (Appendix~\ref{app:isocost}). The cheap judge overtakes our pre-specified FrugalGPT baseline on SGD but only \emph{ties} SAGE-Core there and stays below it on MultiWOZ; the single-call LLM-as-judge in Table~\ref{tab:headline} is a further reduced-cost point SAGE matches or leads. No evaluated judge, at any tested budget, significantly exceeds SAGE-Core, which pays nothing.

\paragraph{Symbolic-injection circularity, and what is non-circular.}
For three classes (WNF, PC, IUV) the injection operator and S1 share a predicate (e.g.\ PC is injected when state writes \texttt{confirmed} with a required field unfilled, exactly the S1 firing condition); where SAGE wins symbolically at near-zero cost (notably S1+S2 on ABCD) it is largely inverting its own injector on the symbolically-decidable subset, and we read those numbers as the compiler \emph{recovering failures of the form it is designed to detect}. UV is closer to non-circular: detection runs an off-the-shelf DeBERTa-v3 NLI step (\texttt{cross-encoder/nli-deberta-v3-base}) that does not share the symbolic predicate, though the injected value is still chosen to be NLI-detectable, so this remains constructive rather than fully in-the-wild. SAGE's UV recall on MultiWOZ (SAGE-Core $0.97$, SAGE-LLM $0.98$) and ABCD ($0.951$/$0.961$) is the strongest external evidence of detection power.

\paragraph{Limitations.}
SAGE requires a workflow specification and is unsuited to open-ended chat. MultiWOZ's partial required-field ordering causes S1 to fire conservatively ($23\%$ FP rate); the dominant SGD failure is value-normalisation mismatch (IUV recall $0.44$); and on ABCD, $5/103$ true failures slip under the recommended SAGE-Core ($4/103$ under SAGE-LLM) when the bot asserts a value the post-turn state does not record. Most fundamentally, all benchmarks use \emph{injected} failures. Our two-annotator human audit (\emph{Human Validation Audit}, Table~\ref{tab:human}; $n{=}200$, $\kappa{=}0.94$) probes the resulting construct-validity risk: it confirms strong fidelity on the transcript-visible classes, but flags WNF as a softer constraint (agreement $0.73$) and IUV as a formal state-consistency signal with weak broad-human salience ($1/30$) rather than a fully human-validated class; all headline conclusions are unchanged when IUV is excluded. A naturally-occurring failure set disjoint from S1's predicates remains the highest-leverage addition and is left to future work. Headline numbers use oracle (service,~intent) routing; learned routers move SAGE $F_1$ by ${<}0.003$ on SGD (Appendix~\ref{app:router}).

\section{Conclusion}
SAGE compiles workflow specifications and per-turn state diffs into atomic criteria, verifies them through a symbolic/encoder cascade with abstention, and returns evidence-bearing turn verdicts. Its recommended \textbf{SAGE-Core} configuration makes zero paid LLM calls and is not significantly exceeded by any evaluated LLM judge across MultiWOZ, SGD, and ABCD slices; \textbf{SAGE-LLM}'s optional paid tier is not load-bearing. A two-annotator human audit ($n{=}200$, $\kappa{=}0.94$) supports strong fidelity for PC and UV (including the ABCD/UV-rich subset) and moderate fidelity for WNF, while identifying IUV as a weaker construct-validity case. These results argue for criterion-level, state-grounded evaluation over holistic transcript scoring for workflow-driven dialogue agents.

\bibliography{sage}

@String(AAAI = {AAAI})

@article{frugalgpt,
  title={{FrugalGPT}: How to Use Large Language Models While Reducing Cost and Improving Performance},
  author={Chen, Lingjiao and Zaharia, Matei and Zou, James},
  journal={Transactions on Machine Learning Research},
  year={2024}
}

@inproceedings{geval,
  title={{G-Eval}: {NLG} Evaluation Using {GPT-4} with Better Human Alignment},
  author={Liu, Yang and Iter, Dan and Xu, Yichong and Wang, Shuohang and Xu, Ruochen and Zhu, Chenguang},
  booktitle={Proceedings of the 2023 Conference on Empirical Methods in Natural Language Processing (EMNLP)},
  year={2023}
}

@inproceedings{mtbench,
  title={Judging {LLM}-as-a-Judge with {MT-Bench} and Chatbot Arena},
  author={Zheng, Lianmin and Chiang, Wei-Lin and Sheng, Ying and Zhuang, Siyuan and Wu, Zhanghao and Zhuang, Yonghao and Lin, Zi and Li, Zhuohan and Li, Dacheng and Xing, Eric P. and Zhang, Hao and Gonzalez, Joseph E. and Stoica, Ion},
  booktitle={Advances in Neural Information Processing Systems},
  year={2023}
}

@inproceedings{multiwoz,
  title={{MultiWOZ} -- A Large-Scale Multi-Domain {Wizard-of-Oz} Dataset for Task-Oriented Dialogue Modelling},
  author={Budzianowski, Pawe{\l} and Wen, Tsung-Hsien and Tseng, Bo-Hsiang and Casanueva, I{\~n}igo and Ultes, Stefan and Ramadan, Osman and Ga{\v{s}}i{\'c}, Milica},
  booktitle={Proceedings of EMNLP},
  year={2018}
}

@inproceedings{multiwoz21,
  title={{MultiWOZ 2.1}: A Consolidated Multi-Domain Dialogue Dataset with State Corrections and State Tracking Baselines},
  author={Eric, Mihail and Goel, Rahul and Paul, Shachi and Sethi, Abhishek and Agarwal, Sanchit and Gao, Shuyang and Kumar, Adarsh and Goyal, Anuj and Ku, Peter and Hakkani-Tur, Dilek},
  booktitle={Proceedings of the 12th Language Resources and Evaluation Conference (LREC)},
  year={2020}
}

@inproceedings{multiwoz24,
  title={{MultiWOZ 2.4}: A Multi-Domain Task-Oriented Dialogue Dataset with Essential Annotation Corrections to Improve State Tracking Evaluation},
  author={Ye, Fanghua and Manotumruksa, Jarana and Yilmaz, Emine},
  booktitle={Proceedings of the 23rd Annual Meeting of the Special Interest Group on Discourse and Dialogue (SIGDIAL)},
  year={2022}
}

@inproceedings{sgd,
  title={Towards Scalable Multi-Domain Conversational Agents: The Schema-Guided Dialogue Dataset},
  author={Rastogi, Abhinav and Zang, Xiaoxue and Sunkara, Srinivas and Gupta, Raghav and Khaitan, Pranav},
  booktitle={Proceedings of the AAAI Conference on Artificial Intelligence},
  year={2020}
}

@inproceedings{sgdx,
  title={{SGD-X}: A Benchmark for Robust Generalization in Schema-Guided Dialogue Systems},
  author={Lee, Harrison and Gupta, Raghav and Rastogi, Abhinav and Cao, Yuan and Zhang, Bin and Wu, Yonghui},
  booktitle={Proceedings of the AAAI Conference on Artificial Intelligence},
  year={2022}
}

@article{summac,
  title={{SummaC}: Re-Visiting {NLI}-Based Models for Inconsistency Detection in Summarization},
  author={Laban, Philippe and Schnabel, Tobias and Bennett, Paul N. and Hearst, Marti A.},
  journal={Transactions of the Association for Computational Linguistics},
  year={2022}
}

@inproceedings{snli,
  title={A Large Annotated Corpus for Learning Natural Language Inference},
  author={Bowman, Samuel R. and Angeli, Gabor and Potts, Christopher and Manning, Christopher D.},
  booktitle={Proceedings of EMNLP},
  year={2015}
}

@inproceedings{mnli,
  title={A Broad-Coverage Challenge Corpus for Sentence Understanding through Inference},
  author={Williams, Adina and Nangia, Nikita and Bowman, Samuel R.},
  booktitle={Proceedings of NAACL},
  year={2018}
}

@inproceedings{deberta,
  title={{DeBERTaV3}: Improving {DeBERTa} using {ELECTRA}-Style Pre-Training with Gradient-Disentangled Embedding Sharing},
  author={He, Pengcheng and Gao, Jianfeng and Chen, Weizhu},
  booktitle={International Conference on Learning Representations},
  year={2023}
}

@inproceedings{sbert,
  title={{Sentence-BERT}: Sentence Embeddings using Siamese {BERT}-Networks},
  author={Reimers, Nils and Gurevych, Iryna},
  booktitle={Proceedings of EMNLP-IJCNLP},
  year={2019}
}

@inproceedings{minilm,
  title={{MiniLM}: Deep Self-Attention Distillation for Task-Agnostic Compression of Pre-Trained Transformers},
  author={Wang, Wenhui and Wei, Furu and Dong, Li and Bao, Hangbo and Yang, Nan and Zhou, Ming},
  booktitle={Advances in Neural Information Processing Systems},
  year={2020}
}

@inproceedings{mpnet,
  title={{MPNet}: Masked and Permuted Pre-training for Language Understanding},
  author={Song, Kaitao and Tan, Xu and Qin, Tao and Lu, Jianfeng and Liu, Tie-Yan},
  booktitle={Advances in Neural Information Processing Systems},
  year={2020}
}

@inproceedings{selective,
  title={Selective Classification for Deep Neural Networks},
  author={Geifman, Yonatan and El-Yaniv, Ran},
  booktitle={Advances in Neural Information Processing Systems},
  year={2017}
}

@inproceedings{cot,
  title={Chain-of-Thought Prompting Elicits Reasoning in Large Language Models},
  author={Wei, Jason and Wang, Xuezhi and Schuurmans, Dale and Bosma, Maarten and Ichter, Brian and Xia, Fei and Chi, Ed H. and Le, Quoc V. and Zhou, Denny},
  booktitle={Advances in Neural Information Processing Systems},
  year={2022}
}

@inproceedings{prometheus,
  title={Prometheus: Inducing Fine-Grained Evaluation Capability in Language Models},
  author={Kim, Seungone and Shin, Jamin and Cho, Yejin and Jang, Joel and Longpre, Shayne and Lee, Hwaran and Yun, Sangdoo and Shin, Seongjin and Kim, Sungdong and Thorne, James and Seo, Minjoon},
  booktitle={International Conference on Learning Representations},
  year={2024}
}

@inproceedings{pandalm,
  title={{PandaLM}: An Automatic Evaluation Benchmark for {LLM} Instruction Tuning Optimization},
  author={Wang, Yidong and Yu, Zhuohao and Yao, Wenjin and Zeng, Zhengran and Yang, Linyi and Wang, Cunxiang and Chen, Hao and Jiang, Chaoya and Xie, Rui and Wang, Jindong and Xie, Xing and Ye, Wei and Zhang, Shikun and Zhang, Yue},
  booktitle={International Conference on Learning Representations},
  year={2024}
}

@article{alpacaeval,
  title={Length-Controlled {AlpacaEval}: A Simple Way to Debias Automatic Evaluators},
  author={Dubois, Yann and Galambosi, Bal{\'a}zs and Liang, Percy and Hashimoto, Tatsunori B.},
  journal={arXiv preprint arXiv:2404.04475},
  year={2024}
}

@inproceedings{routellm,
  title={{RouteLLM}: Learning to Route {LLMs} from Preference Data},
  author={Ong, Isaac and Almahairi, Amjad and Wu, Vincent and Chiang, Wei-Lin and Wu, Tianhao and Gonzalez, Joseph E. and Kadous, M. Waleed and Stoica, Ion},
  booktitle={International Conference on Learning Representations},
  year={2025}
}

@inproceedings{automix,
  title={{AutoMix}: Automatically Mixing Language Models},
  author={Aggarwal, Pranjal and Madaan, Aman and Anand, Ankit and Potharaju, Srividya Pranavi and Mishra, Swaroop and Zhou, Pei and Gupta, Aditya and Rajagopal, Dheeraj and Kappaganthu, Karthik and Yang, Yiming and Upadhyay, Shyam and Faruqui, Manaal and Mausam},
  booktitle={Advances in Neural Information Processing Systems},
  year={2024}
}

@inproceedings{llmblender,
  title={{LLM-Blender}: Ensembling Large Language Models with Pairwise Ranking and Generative Fusion},
  author={Jiang, Dongfu and Ren, Xiang and Lin, Bill Yuchen},
  booktitle={Proceedings of ACL},
  year={2023}
}

@inproceedings{factscore,
  title={{FActScore}: Fine-grained Atomic Evaluation of Factual Precision in Long Form Text Generation},
  author={Min, Sewon and Krishna, Kalpesh and Lyu, Xinxi and Lewis, Mike and Yih, Wen-tau and Koh, Pang Wei and Iyyer, Mohit and Zettlemoyer, Luke and Hajishirzi, Hannaneh},
  booktitle={Proceedings of EMNLP},
  year={2023}
}

@inproceedings{selfcheckgpt,
  title={{SelfCheckGPT}: Zero-Resource Black-Box Hallucination Detection for Generative Large Language Models},
  author={Manakul, Potsawee and Liusie, Adian and Gales, Mark J. F.},
  booktitle={Proceedings of EMNLP},
  year={2023}
}

@inproceedings{ragas,
  title={{RAGAs}: Automated Evaluation of Retrieval Augmented Generation},
  author={Es, Shahul and James, Jithin and Espinosa-Anke, Luis and Schockaert, Steven},
  booktitle={Proceedings of EACL: System Demonstrations},
  year={2024}
}

@inproceedings{abcd,
  title={Action-Based Conversations Dataset: A Corpus for Building More In-Depth Task-Oriented Dialogue Systems},
  author={Chen, Derek and Chen, Howard and Yang, Yi and Lin, Alexander and Yu, Zhou},
  booktitle={Proceedings of NAACL-HLT},
  year={2021}
}

@article{holm,
  title={A Simple Sequentially Rejective Multiple Test Procedure},
  author={Holm, Sture},
  journal={Scandinavian Journal of Statistics},
  volume={6},
  number={2},
  pages={65--70},
  year={1979}
}

@book{efron,
  title={An Introduction to the Bootstrap},
  author={Efron, Bradley and Tibshirani, Robert J.},
  publisher={Chapman \& Hall/CRC},
  year={1993}
}

@article{cohen,
  title={A Coefficient of Agreement for Nominal Scales},
  author={Cohen, Jacob},
  journal={Educational and Psychological Measurement},
  volume={20},
  number={1},
  pages={37--46},
  year={1960}
}

\appendix
\setcounter{secnumdepth}{1} % number appendix sections (A, B, C) so \ref works; body stays unnumbered

\section{Failure Injection Templates}
\label{app:injection}

For each sampled turn we either retain the original (acceptable) turn or programmatically inject one failure from $\mathcal{C}$ by mutating the agent response and the state diff \emph{together}, so the failure is genuinely state-grounded rather than only surface-grounded:
\begin{itemize}
\item \textbf{IUV:} rewrite the post-turn state $s_t$ so that a value $v$ the user supplied in $u_t$ is dropped (the slot remains unfilled or keeps a stale value), forcing a downstream re-ask or stalled workflow.
\item \textbf{WNF:} replace $a_t$ with a request for a slot from another required-field bucket not legal under $\mathrm{Next}(\mathcal{W}, s_{t-1})$.
\item \textbf{PC:} replace $a_t$ with a confirmation surface (``Booking confirmed.'') while leaving $s_{t-1}$ with at least one required slot empty.
\item \textbf{UV:} inject a value into $\Delta_t$ that is plausible for the slot type but does not appear in $h_{<t}\cup\{u_t\}$.
\end{itemize}
The same operator implementations are reused across datasets where applicable; MultiWOZ and SGD use all four classes, while ABCD-UV uses only the UV operator. Only the WorkflowSpec source differs.

\section{Full Per-Method Result Tables}
\label{app:fulltables}

\begin{figure}[!t]
\centering
\includegraphics[width=0.95\columnwidth]{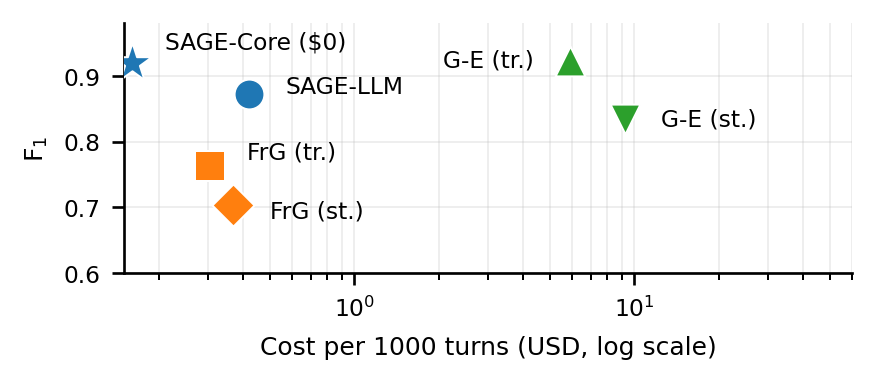}
\caption{Cost-quality Pareto on MultiWOZ~2.4 UV-rich, all six Table~\ref{tab:uvrich} configurations (log-cost axis; SAGE-Core's \$0.00 is clamped to the left edge). \textbf{SAGE-Core} sits at the Pareto corner -- essentially tied on $F_1$ with the top method at zero paid API cost. Transcript-only G-Eval is the top-$F_1$ point but at \$4.70/1k and $12.0$s median latency (vs.\ SAGE-Core's \$0 paid API cost and $0.03$s warm local latency; $\approx 14\times$ the cost and $\approx 44\times$ the latency relative to SAGE-LLM); SAGE-LLM (adding S3) is dominated by SAGE-Core here. State-aware variants of both LLM judges fall below their transcript-only counterparts.}
\label{fig:cost_quality}
\end{figure}

Tables~\ref{tab:mwabcd},~\ref{tab:uvrich}, and~\ref{tab:sgd} give the per-method precision, recall, $F_1$, dollar cost, and latency underlying the headline summary (Table~\ref{tab:headline}); Figure~\ref{fig:cost_quality} plots the UV-rich cost-quality frontier.

\begin{table}[!t]
\centering
\footnotesize
\setlength{\tabcolsep}{2.2pt}
\renewcommand{\arraystretch}{1.15}
\caption{Headline cost-quality on two slices. \textbf{MW-mixed}: MultiWOZ~2.4, all four failure classes mixed ($n{=}250$, $126$ injected). \textbf{ABCD-UV}: Action-Based Conversations Dataset~\citep{abcd}, UV-only injection ($n{=}223$, $103$ injected, $30$ workflows). \textbf{SAGE-Core}=compiler+S1+S2 (zero paid LLM, recommended); \textbf{SAGE-LLM}=SAGE-Core+optional S3. LLM-judge is the stronger-mode single-call \texttt{gpt-4.1} verdict per slice (state-aware on MW-mixed; transcript-only on ABCD). Dollar cost is measured tokens at OpenAI API list prices; SAGE-Core makes no API call, so its cost is \$0.00 (its symbolic/on-device compute is not monetized). Best per column in \textbf{bold}. Paired bootstrap of SAGE-Core vs.\ the strongest per-slice baseline: MW-mixed $\Delta F_1{=}{+}.069$ ($p{=}.022$), ABCD-UV ${+}.056$ ($p{=}.044$). SAGE-Core $F_1$ $95\%$ bootstrap CI -- MW-mixed $[.818,.909]$, ABCD-UV $[.889,.962]$; SAGE-LLM -- MW-mixed $[.781,.879]$, ABCD-UV $[.841,.929]$; LLM-judge $[.745,.845]$/$[.823,.916]$, G-Eval $[.711,.835]$/$[.726,.852]$, FrugalGPT $[.683,.791]$/$[.596,.718]$.}
\label{tab:mwabcd}
\begin{tabular}{@{}lcccc@{\hspace{6pt}}cccc@{}}
\hline
& \multicolumn{4}{c}{MW-mixed} & \multicolumn{4}{c}{ABCD-UV} \\
\cline{2-5}\cline{6-9}\\[-2.0ex]
Method & $F_1$ & P & R & \$\,/1k & $F_1$ & P & R & \$\,/1k \\
\hline
SAGE-Core    & \textbf{.866} & .884          & .849          & \textbf{.00} & \textbf{.929} & \textbf{.907} & .951           & \textbf{.00} \\
SAGE-LLM     & .833          & .797          & .873          & .46          & .888          & .825          & .961           & 1.37 \\
LLM-judge    & .797          & .694          & .937          & 1.74         & .873          & .821          & .932           & 1.52 \\
G-Eval (tr.) & .777          & \textbf{.965} & .651          & 4.68         & .794          & .823          & .767           & 5.93 \\
FrugalGPT    & .740          & .590          & \textbf{.992} & 4.63         & .660          & .493          & \textbf{1.00}  & 1.48 \\
\hline
\end{tabular}
\end{table}

\begin{table}[!t]
\centering
\footnotesize
\setlength{\tabcolsep}{3pt}
\renewcommand{\arraystretch}{1.15}
\caption{Single-class UV stress test on MultiWOZ~2.4 ($n{=}250$, $121$ injected \texttt{unsupported\_value} failures). \textbf{SAGE-Core}=compiler+S1+S2 (zero paid LLM); \textbf{SAGE-LLM}=SAGE-Core+optional S3. Cost from measured tokens and OpenAI API list prices; latency wall-clock per-turn (SAGE-Core issues no API call, so its p50 is warm in-process S1+S2 compute, not directly comparable to the API-based rows' harness wall-clock). \textbf{tr.}=transcript-only, \textbf{st.}=state-aware. Best per column in \textbf{bold}. Paired bootstrap SAGE-Core vs.\ G-Eval-tr (the strongest baseline here): $\Delta F_1{=}{-}.003$, $95\%$ CI $[{-}.052,{+}.046]$, $p{=}.89$ (tie). $F_1$ $95\%$ bootstrap CI: SAGE-Core $[.883,.952]$, SAGE-LLM $[.828,.913]$, FrugalGPT-tr $[.707,.813]$, FrugalGPT-st $[.645,.756]$, G-Eval-tr $[.885,.955]$, G-Eval-st $[.776,.885]$.}
\label{tab:uvrich}
\begin{tabular}{@{}lcccccr@{}}
\hline
Method & P & R & $F_1$ & Acc. & \$\,/\,1k & p50 (s) \\
\hline
SAGE-Core         & 0.891          & 0.950          & 0.920          & 0.920          & \textbf{0.00} & \textbf{0.03} \\
SAGE-LLM          & 0.796          & 0.967          & 0.873          & 0.864          & 0.34          & 0.27 \\
FrugalGPT (tr.)   & 0.617          & \textbf{1.000} & 0.763          & 0.700          & 0.82          & 4.93 \\
FrugalGPT (st.)   & 0.548          & 0.983          & 0.704          & 0.600          & 0.94          & 4.21 \\
G-Eval (tr.)      & 0.956          & 0.893          & \textbf{0.923} & \textbf{0.928} & 4.70          & 12.0 \\
G-Eval (st.)      & \textbf{0.978} & 0.727          & 0.834          & 0.860          & 7.39          & 12.0 \\
\hline
\end{tabular}
\end{table}

\begin{table}[!t]
\centering
\footnotesize
\setlength{\tabcolsep}{3pt}
\renewcommand{\arraystretch}{1.15}
\caption{Cross-dataset on Schema-Guided Dialogue~\citep{sgd} ($n{=}280$, $151$ injected failures, $37$ (service, intent) workflows). Same evaluator code as Table~\ref{tab:uvrich}; only the WorkflowSpec source differs. \textbf{SAGE-Core}=zero-paid-LLM S1+S2; \textbf{SAGE-LLM}=+optional S3 (latency/cost as Table~\ref{tab:uvrich}). \textbf{tr.}=transcript-only, \textbf{st.}=state-aware. Paired bootstrap SAGE-Core vs.\ the strongest evaluated GPT-4.1 baseline (FrugalGPT-tr on the full slice), re-selecting the maximum within each resample: $\Delta F_1{=}{+}.075$, $95\%$ CI $[{+}.015,{+}.102]$, $p{=}.011$. $F_1$ $95\%$ bootstrap CI: SAGE-Core $[.820,.904]$, SAGE-LLM $[.804,.891]$, FrugalGPT-tr $[.738,.834]$, FrugalGPT-st $[.726,.821]$, G-Eval-tr $[.725,.837]$, G-Eval-st $[.724,.837]$.}
\label{tab:sgd}
\begin{tabular}{@{}lcccccr@{}}
\hline
Method & P & R & $F_1$ & Acc. & \$\,/\,1k & p50 (s) \\
\hline
SAGE-Core         & 0.888          & 0.841          & \textbf{0.864} & \textbf{0.857} & \textbf{0.00} & \textbf{0.03} \\
SAGE-LLM          & 0.853          & 0.848          & 0.850          & 0.839          & 0.44          & 0.88 \\
FrugalGPT (tr.)   & 0.693          & 0.914          & 0.789          & 0.736          & 0.83          & 4.61 \\
FrugalGPT (st.)   & 0.645          & \textbf{0.974} & 0.776          & 0.696          & 0.99          & 3.69 \\
G-Eval (tr.)      & 0.962          & 0.662          & 0.784          & 0.804          & 5.10          & 12.0 \\
G-Eval (st.)      & \textbf{0.990} & 0.649          & 0.784          & 0.807          & 8.02          & 12.0 \\
\hline
\end{tabular}
\end{table}

\begin{figure}[!t]
\centering
\includegraphics[width=0.95\columnwidth]{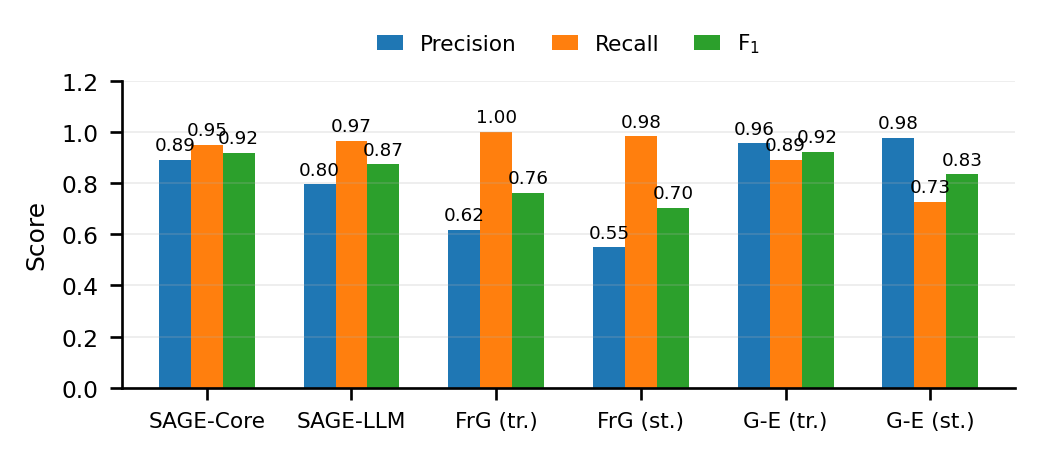}
\caption{Precision, recall, and $F_1$ on MultiWOZ~2.4 UV-rich ($n{=}250$); exact values in Table~\ref{tab:uvrich}. SAGE-Core (zero paid LLM) attains the second-highest $F_1$, statistically tied with transcript-only G-Eval, which holds the top $F_1$ on this single-class slice but at \$4.70/1k and $12.0$s median latency (vs.\ SAGE-Core's \$0 and $0.03$s; $\approx 14\times$ cost and $\approx 44\times$ latency relative to SAGE-LLM); enabling S3 (SAGE-LLM) does not help here.}
\label{fig:pr_f1}
\end{figure}

\section{Router Robustness Details}
\label{app:router}

This appendix expands the ``Oracle routing'' paragraph in the Discussion. We measured router robustness on the SGD cross-dataset slice ($n{=}296$, headline injectors). Two zero-shot routers were evaluated end-to-end against the gold (service, intent) oracle used in the headline tables:

\paragraph{MPNet bi-encoder.} A pretrained MPNet~\citep{mpnet} bi-encoder (\texttt{sentence-transformers/all-mpnet-base-v2}, revision \texttt{e8c3b32}) encodes the schema's service+intent natural-language descriptions and the dialogue prefix; cosine similarity selects one of the $37$ (service, intent) candidates. Top-1 accuracy on the SGD slice is $0.476$; top-8 accuracy is $0.892$.

\paragraph{LLM-on-shortlist.} \texttt{gpt-4.1-mini} is prompted with the MPNet top-$8$ candidates (rendered as schema headers) and the dialogue prefix, and constrained to pick one. Top-1 accuracy is $0.804$. This pattern is a cheap distillation: the bi-encoder narrows the search space; the LLM disambiguates the residual.

\paragraph{End-to-end results.} We report the recommended \textbf{SAGE-Core} (S1+S2, \texttt{no\_S3}) configuration; it issues no S3 calls and is therefore deterministic on fixed turns (no regeneration band). $F_1$ on this $n{=}296$ SGD slice:
\begin{itemize}
\item Oracle routing: $F_1 = 0.861$
\item LLM-on-shortlist router: $F_1 = 0.861$
\item MPNet router: $F_1 = 0.861$
\end{itemize}
All three coincide to within $<0.001$ ($P{=}0.887$, $R{=}0.836$ throughout): the router choice does not move SAGE-Core's $F_1$ on this slice. On the $155$ MPNet-misrouted turns specifically SAGE-Core still reaches $F_1 = 0.862$; on the $58$ LLM-router-misrouted turns it reaches $0.836$. For reference, the full-cascade \textbf{SAGE-LLM} exhibits the same invariance at nonzero API cost ($0.854 / 0.852 / 0.855$ for oracle/LLM/MPNet). These $n{=}296$ numbers are not directly comparable to the $n{=}280$ headline SGD $F_1$ (SAGE-Core $0.864$, Table~\ref{tab:sgd}); the slice differs.

\paragraph{Why misrouting barely hurts.} SGD intent confusions are overwhelmingly intra-service: services like \texttt{Restaurants\_1} and \texttt{Restaurants\_2} share most slot names; \texttt{Travel\_1} confusions are between sub-intents (\texttt{FindAttractions} vs.\ \texttt{GetAttractionInfo}) with overlapping slot spaces. A misrouted scenario therefore shares the slot space of the correct one, so the UV grounding check (values must be entailed by history) and most S1 transition rules still fire on the right signals. SAGE's quality is governed by the per-criterion verifier cascade rather than by workflow selection: in our tested SGD setting, even the MPNet router at $0.476$ top-1 accuracy produced no measurable SAGE-Core $F_1$ degradation, because most routing errors stayed within services with overlapping slot spaces.

\paragraph{Implication for the head-to-head.} Because oracle routing introduces $<0.003$ $F_1$ bias on the SAGE side, and the transcript-only baselines do not consume schema at all (state-aware variants consume the same schema/state context SAGE sees but do not route per-criterion), the SAGE/G-Eval/FrugalGPT comparison in Tables~\ref{tab:uvrich}--\ref{tab:sgd} is not confounded by the oracle. Removing the oracle from SAGE and adding a router to the baselines would not change the relative ranking.

\section{Cheaper-Backbone Judge Comparison}
\label{app:isocost}

A natural objection to the headline cost story is that SAGE's advantage might be a \emph{model-class} artifact: the holistic baselines use GPT-4.1, and a cheaper backbone would close the cost gap without changing the ranking. To test this we re-ran G-Eval on the $\approx 5\times$ cheaper (at matched token usage) \texttt{gpt-4.1-mini} backbone, in both state-aware and transcript-only prompt variants, on the identical MultiWOZ UV-rich and SGD headline slices (Table~\ref{tab:isocost}; SAGE rows use the headline-slice per-turn verdicts, not the regenerated ablation slice, so all rows in the table are on one generation). Swapping the backbone cuts the holistic judge's cost to \$0.9--1.6/1k, versus \$4.7--8.0/1k for the GPT-4.1 judge (depending on prompt mode and slice; $\approx 5\times$ cheaper at matched token usage), and it materially changes the baseline ranking on SGD: the state-aware mini judge reaches $0.859$, overtaking our pre-specified strongest SGD baseline (FrugalGPT-tr, $0.789$). It is therefore the strongest SGD judge in our evaluation. Even so, a paired bootstrap (SAGE-Core vs.\ this judge, same turns) shows only a statistical \emph{tie}, not a loss: SGD $\Delta F_1{=}{+}.005$ ($0.864$ vs.\ $0.859$, $95\%$ CI $[{-}.045,{+}.054]$, $p{=}.83$); on MultiWOZ the best cheap judge ($0.901$, transcript-only) stays below SAGE-Core ($0.920$; $\Delta F_1{=}{+}.019$, $[{-}.033,{+}.071]$, $p{=}.46$). These two reduced-cost paired tests are uncorrected, but both are ties, so multiplicity does not affect the conclusion. No reduced-cost judge \emph{exceeds} SAGE-Core on either slice, while SAGE-Core issues \emph{zero} paid API calls.

To keep the ``no judge on a cheaper backbone exceeds SAGE-Core'' claim load-bearing on \emph{all four} headline slices rather than only the two single-class stress slices in Table~\ref{tab:isocost}, we also ran both mini variants on the two multi-class slices (MW-mixed and ABCD-UV). There the gap widens rather than narrows: the best reduced-cost mini judge reaches only $0.759$ on MW-mixed (transcript-only) and $0.777$ on ABCD (state-aware), well below SAGE-Core's $0.866$ and $0.929$. This holds even on MW-mixed, the slice where state context \emph{helped} the GPT-4.1 judge ($+0.063$); the state-aware mini judge there scores just $0.698$. The paired bootstrap confirms SAGE-Core wins significantly on both (MW-mixed $\Delta F_1{=}{+}.107$, $p{=}.001$; ABCD $\Delta F_1{=}{+}.152$, $p{<}.001$). Summarising across all four slices: no reduced-cost mini judge exceeds SAGE-Core anywhere -- it ties the strongest mini judge on the two single-class stress slices (UV-rich, SGD) and significantly beats it on the two multi-class slices, at \$0.00.

The cheaper backbone also flips the state-context effect: the mini judge is \emph{helped} by state context on SGD ($0.789\!\to\!0.859$, ${+}0.070$), the reverse of the GPT-4.1 judges, which are \emph{hurt} by it (up to $-0.09$). This scopes our ``state context lowers $F_1$'' finding to the GPT-4.1 prompt implementation rather than to holistic judges in general, and correspondingly narrows the ``architectural, not informational'' argument (Discussion): the defensible claim is that even the best-informed, best-backboned holistic judge does not significantly exceed SAGE-Core. Because SAGE-Core (S1+S2) also matches or beats the full SAGE-LLM cascade, the $F_1$ gap is \emph{not merely a backbone artifact} and is consistent with the architectural benefit of state-grounded criterion decomposition rather than the choice of LLM backbone.

\begin{table}[t]
\centering
\footnotesize
\setlength{\tabcolsep}{3pt}
\renewcommand{\arraystretch}{1.15}
\caption{\textbf{Reduced-cost LLM-judge comparison.} Replacing G-Eval's GPT-4.1 backbone with the $\approx 5\times$ cheaper (at matched token usage) GPT-4.1-mini (in both state-aware \textbf{st.} and transcript-only \textbf{tr.} prompt variants) cuts the holistic judge's cost to \$0.9--1.6/1k, versus \$4.7--8.0/1k for GPT-4.1, yet no reduced-cost judge exceeds SAGE-Core's $F_1$ on either slice: the best cheap judge reaches $0.901$ on MultiWOZ (vs.\ SAGE-Core $0.920$) and $0.859$ on SGD (vs.\ $0.864$), while SAGE-Core issues \emph{zero} paid API calls. On SGD the state-aware mini judge ($0.859$) overtakes the pre-specified FrugalGPT baseline but only \emph{ties} SAGE-Core (paired bootstrap $p{=}.83$); no reduced-cost judge \emph{exceeds} SAGE-Core on either slice, so the gap is \emph{not merely a backbone artifact} and is consistent with SAGE's state-grounded architecture. p50 (s) is the per-turn headline-harness wall time (consistent with Tables~\ref{tab:uvrich}--\ref{tab:sgd}); SAGE rows are warm in-process compute and are not directly comparable to the API rows' harness wall-clock.}
\label{tab:isocost}
\begin{tabular}{@{}lcccccc@{}}
\hline
& \multicolumn{3}{c}{MultiWOZ} & \multicolumn{3}{c}{SGD} \\
Evaluator & $F_1$ & \$/1k & p50\,s & $F_1$ & \$/1k & p50\,s \\
\hline
SAGE-Core (S1+S2) & 0.920 & 0.00 & 0.03 & 0.864 & 0.00 & 0.03 \\
SAGE-LLM (full) & 0.873 & 0.34 & 0.27 & 0.850 & 0.44 & 0.88 \\
G-Eval (4.1, st.) & 0.834 & 7.39 & 11.99 & 0.784 & 8.02 & 11.98 \\
G-Eval (4.1-mini, st.) & 0.883 & 1.47 & 7.19 & 0.859 & 1.60 & 7.18 \\
G-Eval (4.1-mini, tr.) & 0.901 & 0.93 & 7.20 & 0.789 & 1.01 & 7.20 \\
FrugalGPT (tr.) & 0.763 & 0.82 & 4.93 & 0.789 & 0.83 & 4.61 \\
\hline
\end{tabular}
\end{table}

\section{Structured Repair Signals}
\label{app:repair}

Every SAGE verdict carries an evidence trace, and for a detected failure the aggregator emits a \emph{repair signal}: the concrete state edit or clarifying action that would resolve the violated criterion. These are a side output of the per-criterion decomposition -- we do not evaluate a repair-execution loop -- but they illustrate what the evidence trace makes actionable. Table~\ref{tab:repair} shows representative examples drawn from the MultiWOZ and SGD traces, one per failure class.

\begin{table}[t]
\centering
\footnotesize
\setlength{\tabcolsep}{4pt}
\renewcommand{\arraystretch}{1.25}
\caption{Representative structured repair signals emitted alongside SAGE evidence traces (illustrative side output; not part of the quantitative evaluation). Class codes: IUV = ignored user value, PC = premature confirmation, UV = unsupported value.}
\label{tab:repair}
\begin{tabular}{@{}p{0.06\columnwidth}p{0.44\columnwidth}p{0.38\columnwidth}@{}}
\hline
Class & Evidence trace & Repair signal \\
\hline
IUV & User says ``2 people,'' but $\Delta_t$ omits \texttt{book\_people} & \texttt{write book\_people=2} \\
PC & Agent confirms while a required slot is still unfilled ($\mathcal{O}(s_{t-1})$ false) & Ask the missing field before confirming \\
UV & $\Delta_t$ writes \texttt{destination} not entailed by $h_{<t}\cup\{u_t\}$ & Remove the value or ask the user to clarify \\
\hline
\end{tabular}
\end{table}

\section{Per-Tier Ablation}
\label{app:ablation}

Table~\ref{tab:ablation} reports per-tier leave-one-out ablations. S1 is the only tier compiled for WNF, PC, and IUV; on SGD, disabling S1 drops recall $0.848\!\to\!0.682$, with WNF/IUV collapsing to zero. On MultiWOZ, by contrast, removing S1 \emph{raises} $F_1$ ($0.870\!\to\!0.911$): the partial slot ordering in MultiWOZ makes some S1 next-field checks fire conservatively on legal turns, so dropping S1 removes those false positives; on SGD, whose schema ordering is cleaner, S1 remains necessary for WNF/IUV recall. S2 carries the precision floor: removing it drops precision $0.87\!\to\!0.66$ on SGD and $0.79\!\to\!0.61$ on MultiWOZ. On both injected slices the \textbf{SAGE-Core} (S1+S2, \texttt{no\_S3}) configuration is stronger than \textbf{SAGE-LLM} (full) -- this is the deployed operating point; a pinned-config rerun of ABCD confirms the same pattern (SAGE-LLM $F_1{=}0.888 \to$ SAGE-Core $0.929$, the latter at zero LLM cost). S3 is retained as an extensibility hook for open-class failures not expressible in the compiled criteria; on the injected slices in this paper it does not net positive, and we do not claim it as a load-bearing contribution.

\begin{table}[!t]
\centering
\footnotesize
\caption{Per-tier leave-one-out ablation on both benchmark slices. $F_1$ over the full set of original turns; orphaned criteria abstain and any gold failure they would catch is a miss. Latency is per-turn wall time on the current dev box; absolute values are higher than Table~\ref{tab:uvrich} because the headline run used a faster machine. This ablation was run on an independently regenerated slice (same sampled turns, pipeline re-run), so absolute $F_1$ can differ from Tables~\ref{tab:headline}--\ref{tab:sgd} by up to $0.006$ (the run-to-run regeneration band, e.g.\ SGD \texttt{full} $0.856$ here vs.\ $0.850$ in Table~\ref{tab:sgd}); this band comes entirely from S3/API stochasticity in the \texttt{full} config, whereas SAGE-Core (\texttt{no\_S3}) is bitwise deterministic on fixed turns and its P/R/$F_1$ are therefore identical across generations. The ablation should be read for within-table tier contributions rather than absolute cross-table comparison. Best per slice in bold. \texttt{full}=\textbf{SAGE-LLM} (S1+S2+S3, the full-cascade config); \texttt{no\_S3}=\textbf{SAGE-Core} (S1+S2), the recommended zero-paid-LLM operating point for compiled taxonomies, which matches or exceeds SAGE-LLM on both slices at \$0.00. $F_1$ $95\%$ bootstrap CIs (MW$/$SGD): full $[.82,.91]/[.81,.90]$, no\_S1 $[.87,.94]/[.74,.85]$, no\_S2 $[.68,.79]/[.67,.78]$, no\_S3 $[.88,.95]/[.82,.90]$.}
\label{tab:ablation}
\setlength{\tabcolsep}{4pt}
\renewcommand{\arraystretch}{1.15}
\begin{tabular}{@{}llccccr@{}}
\hline
Dataset & Config & P & R & $F_1$ & \$\,/\,1k & p50 (s) \\
\hline
MultiWOZ & full & 0.791 & 0.967 & 0.870 & 0.34 & 1.00 \\
 & no\_S1 & 0.860 & 0.967 & 0.911 & 0.34 & 0.93 \\
 & no\_S2 & 0.608 & 0.934 & 0.736 & 1.44 & 1.75 \\
 & no\_S3 & 0.891 & 0.950 & \textbf{0.920} & 0.00 & 0.22 \\
\hline
SGD & full & 0.865 & 0.848 & 0.856 & 0.44 & 0.74 \\
 & no\_S1 & 0.963 & 0.682 & 0.798 & 0.44 & 0.92 \\
 & no\_S2 & 0.659 & 0.808 & 0.726 & 1.52 & 1.55 \\
 & no\_S3 & 0.888 & 0.841 & \textbf{0.864} & 0.00 & 0.17 \\
\hline
\end{tabular}
\end{table}

\section{Cascade Routing and Cost Anatomy}
\label{app:routing}

Table~\ref{tab:routing} reports how SAGE-LLM's cascade resolves criteria, making the ``abstention-aware'' mechanism concrete (under SAGE-Core the S3-routed criteria abstain instead of being decided by a paid LLM call; Table~\ref{tab:selective} reports the resulting undecided rates). Across all four slices the majority of criteria are decided at zero API cost: $56$--$65\%$ by S1's symbolic rules and a further $25$--$31\%$ by S2's on-device encoder, leaving $9$--$19\%$ that escalate to the focused-LLM stage S3. Between $24\%$ and $53\%$ of \emph{turns} invoke S3 at all, giving $0.24$--$0.53$ paid LLM calls per turn for SAGE-LLM versus the one-or-more full-transcript calls every holistic judge pays; SAGE-Core makes zero. On the multi-criteria MultiWOZ/SGD slices this escalation profile makes SAGE-LLM roughly $12$--$22\times$ cheaper than the GPT-4.1 G-Eval judge (depending on prompt mode and slice), and SAGE-Core removes even that residual cost. On the UV-only ABCD slice a larger share of criteria are open-class and reach S3, so SAGE-LLM is only marginally cheaper than the single-call judge ($\$1.37$ vs.\ $\$1.52$/1k); SAGE-Core decides ABCD at no API cost and \emph{higher} $F_1$ ($0.929$ vs.\ $0.888$), which is why we recommend it.

\begin{table}[!t]
\centering
\footnotesize
\setlength{\tabcolsep}{4pt}
\renewcommand{\arraystretch}{1.15}
\caption{Cascade routing per slice: share of compiled criteria resolved at each stage (S1 symbolic / S2 encoder-NLI / S3 focused-LLM), fraction of turns with a criterion resolved at S3, and mean paid LLM calls per turn. S1 and S2 incur zero API cost; only S3 calls are billed.}
\label{tab:routing}
\begin{tabular}{@{}lccccc@{}}
\hline
Slice & S1\,\% & S2\,\% & S3\,\% & Turns\,$\to$\,S3 & LLM/turn \\
\hline
MW-mixed & 65.3 & 24.6 & 10.0 & 33.2\% & 0.332 \\
UV-rich  & 61.1 & 30.2 & \phantom{0}8.7 & 24.4\% & 0.244 \\
SGD      & 57.2 & 30.9 & 11.9 & 30.0\% & 0.300 \\
ABCD-UV  & 55.8 & 25.4 & 18.9 & 53.4\% & 0.534 \\
\hline
\end{tabular}
\end{table}

\section{Selective Evaluation}
\label{app:selective}

The confidence gate that routes low-margin S2 decisions upward is also a turn-level abstention signal: assigning each turn the minimum decision margin across its decided criteria (deterministic S1 decisions score maximal margin) yields a ranking whose lowest-confidence tail can be declined. Table~\ref{tab:selective} reports these numbers on the \textbf{SAGE-Core (S1+S2, \texttt{no\_S3}) operating point} -- the recommended deployed configuration, which is also the SAGE-Core headline rows of Tables~\ref{tab:headline}--\ref{tab:sgd} (\emph{not} the SAGE-LLM rows). These $F_1@100\%$ values are computed on the identical per-turn slices used for every baseline (same turn IDs and gold labels, verified row-for-row); because the S1+S2 pipeline is deterministic given the compiled plan, they coincide with the deterministic \texttt{no\_S3} leave-one-out ablation (Table~\ref{tab:ablation}) to within that ablation's stated regeneration band, and for ABCD the reproduction anchor is the pinned-config S1+S2 rerun ($0.929$). We report the full-coverage $F_1$ (undecided\,$\to$\,acceptable), the all-abstain \emph{undecided} rate, and selective $F_1$ when the least-confident $10\%$/$20\%$ of turns are declined. Declining the least-confident decile raises $F_1$ on the three multi-margin slices (MW-mixed ${+}0.023$, SGD ${+}0.009$, UV-rich ${+}0.008$): the signal is \emph{calibrated} -- the turns SAGE is least sure about are disproportionately the ones it gets wrong -- so an operator can trade coverage for reliability along a risk--coverage frontier rather than being forced into a single operating point. ABCD behaves differently and instructively: its $10.8\%$ undecided turns are \emph{all} gold-acceptable (so full-coverage scoring is lossless there and $F_1@90\%$ is unchanged at $0.929$), while its residual $F_1$ errors are \emph{confident} symbolic S1 decisions rather than low-margin S2 ones -- so declining the low-confidence tail cannot reorder them and $F_1$ is flat-to-slightly-negative (${-}0.009$ at $80\%$ coverage). Selective evaluation thus buys reliability precisely when the residual errors live in the low-margin region (the encoder-heavy slices), and correctly abstains-without-improving when they are confident symbolic decisions (ABCD). This realises the abstention-aware aggregator as a tunable selective evaluator, not merely a cost-routing device.

\begin{table}[!t]
\centering
\footnotesize
\setlength{\tabcolsep}{4pt}
\renewcommand{\arraystretch}{1.15}
\caption{Selective evaluation of the \textbf{SAGE-Core (S1+S2, \texttt{no\_S3}) operating point} -- \emph{not} the SAGE-LLM headline config (Tables~\ref{tab:headline}--\ref{tab:sgd}). \textbf{Undec.}=fraction of turns where every compiled criterion abstains (mapped to acceptable under full coverage). $F_1@c$=$F_1$ on the most-confident $c$ of turns, ranked by minimum per-criterion decision margin. $F_1@100\%$ is computed on the identical per-turn slices as all baselines (verified row-for-row); given the deterministic S1+S2 pipeline it coincides with the \texttt{no\_S3} rows of Table~\ref{tab:ablation} (MW/SGD) within that ablation's regeneration band, and matches the pinned S1+S2 ABCD rerun. Declining the least-confident decile improves $F_1$ on the three encoder-heavy slices; ABCD's undecided turns are all acceptable and its residual errors are confident symbolic decisions, so its curve is flat-to-slightly-negative.}
\label{tab:selective}
\begin{tabular}{@{}lcccc@{}}
\hline
Slice & Undec. & $F_1@100\%$ & $F_1@90\%$ & $F_1@80\%$ \\
\hline
MW-mixed & 2.8\% & 0.866 & 0.889 & 0.878 \\
UV-rich  & 1.6\% & 0.920 & 0.928 & 0.928 \\
SGD      & 6.4\% & 0.864 & 0.873 & 0.876 \\
ABCD-UV  & 10.8\% & 0.929 & 0.929 & 0.920 \\
\hline
\end{tabular}
\end{table}

\section{Schema-Paraphrase Robustness (SGD-X)}
\label{app:sgdx}

To test whether SAGE-Core's $F_1$ rests on schema-grounded reasoning rather than canonical-name overlap, we replay the recommended \textbf{SAGE-Core} (S1+S2, \texttt{no\_S3}) configuration on SGD-X~\citep{sgdx} ($5$ paraphrased schema variants, same $n{=}280$ slice). Mean $F_1$ ($\sigma$, invariant-verdict share): SAGE-Core $0.628$ ($0.002$, $99.3\%$), FrugalGPT $0.587$ ($0.005$, $74.3\%$), G-Eval $0.498$ ($0.015$, $93.2\%$). All three drop relative to canonical SGD, but G-Eval's drop is largest ($0.784\!\to\!0.498$); SAGE-Core has $\approx 37\times$ fewer cross-variant disagreements than FrugalGPT and $\approx 10\times$ fewer than G-Eval. SAGE-Core's own absolute drop ($0.864\!\to\!0.628$, relative to its canonical SGD $F_1$ in Table~\ref{tab:sgd}) comes almost entirely from S2: paraphrased slot names weaken the NLI encoder's entailment margins on UV/IUV criteria, lowering recall, while S1's symbolic checks -- keyed to slot \emph{identity} rather than surface name -- stay stable. The claim is thus low \emph{variance} across paraphrases ($\sigma{=}0.002$), not paraphrase-immune accuracy.

\section{Reproducibility Details}
\label{app:repro}

\paragraph{Hyperparameter values tried and selection criterion.}
SAGE has two tunable numbers, both in S2. The entailment margin threshold $\tau$ on $m = p(\text{ent}) - p(\text{contra})$ was selected as $\tau{=}0.20$ by maximising $F_1$ on the $40$-turn development split disjoint from every evaluation slice, and then applied unchanged to MultiWOZ, SGD, and ABCD. As a post-hoc robustness check we additionally swept $\tau$ over seven values -- $\{0.05, 0.10, 0.15, 0.20, 0.25, 0.30, 0.40\}$ -- on the full UV-rich slice ($n{=}250$), the slice that most exercises the entailment path. $F_1$ varied only within $[0.881, 0.888]$ (a range of $0.007$, with precision $0.810$--$0.822$ and recall constant at $0.965$), and cost varied within $\$0.33$--$\$0.37$ per $1$k turns as the abstention rate shifted the S3 escalation volume. The reported operating point is therefore not a tuned optimum: any threshold in the swept range reproduces the headline conclusion. The extraction-confidence floor was fixed at $0.5$ a priori and not swept. Baseline thresholds (the G-Eval-style judge's cut on the $1$--$5$ acceptability scale and the FrugalGPT-style cascade's escalation confidence) were likewise fixed on the development split and held constant across slices.

\paragraph{Number of runs per reported result.}
Every evaluator is executed exactly once per slice. This is the correct protocol here because all components are deterministic under the settings above: the symbolic compiler and S1 are pure functions of the schema and state diff, the S2 encoders are deterministic at pinned revisions, and every LLM call (S3 and all three baseline families) runs at temperature $0$ against a pinned model snapshot. Reported uncertainty therefore comes from resampling turns -- $10{,}000$ bootstrap replicates at seed $0$ -- rather than from repeated runs. The single exception is the G-Eval-style judge, which draws $n{=}6$ samples per turn \emph{within} its single run by construction; those samples are aggregated into one score before scoring, so it too contributes one verdict per turn.

\paragraph{Computing infrastructure.}
All symbolic and encoder computation ran on a single laptop-class CPU: an Intel Core Ultra 7 268V ($8$ cores, $8$ threads) with $31.7$\,GB RAM under Windows~11 Enterprise (build $10.0.26100$). No GPU was used; the S2 encoders and the MPNet router run on CPU under a CPU-only PyTorch build, which is why SAGE-Core's warm local p50 latency of $0.03$\,s is achievable without accelerator hardware and why we report its compute as unmonetized. The software stack is Python $3.11.9$, PyTorch $2.12.0$ (CPU build), Transformers $5.8.1$, and Sentence-Transformers $5.5.1$. S3 and all LLM baselines call Azure-hosted \texttt{gpt-4.1}/\texttt{gpt-4.1-mini} (snapshot \texttt{2025-04-14}) over the network, so their latencies include API round-trip time and are not CPU-bound.

\section{Per-Evaluator IUV Detection}
\label{app:iuv}

On the $30$ audited IUV turns, detection trades sharply against precision: transcript-only LLM judges catch $\le 2/30$, a state-aware LLM judge $7/30$, a reduced-cost state-aware GPT-4.1-mini judge $13/30$ (at a $9\%$ clean-turn flag rate), SAGE-Core $17/30$ at an $8\%$ clean-turn flag rate, and a trigger-happy baseline $27/30$ only by flagging $70\%$ of clean turns. No evaluator dominates; SAGE-Core sits at the high-precision end of this frontier, which is why we report it as the recommended operating point rather than claiming it uniquely detects IUV.

\end{document}